\documentclass{elsarticle}

\usepackage{lineno}

\usepackage[hyphens]{url}
\usepackage{multirow}
\usepackage{float}
\usepackage{array}
\usepackage{booktabs}

\usepackage{graphicx}

\usepackage{siunitx}

\usepackage{color}
\usepackage{colortbl}
\usepackage{epstopdf}
\usepackage{amssymb}
\usepackage{amsmath}
\usepackage{subcaption}

\usepackage{multicol}

\modulolinenumbers[5]

\journal{Journal of \LaTeX\ Templates}

\usepackage{algorithm} % algorithm floating environment
\floatname{algorithm}{Algorithm}

\newcommand{\INDSTATE}[1][1]{\hspace{0.25 cm}}

\usepackage{algcompatible}

\algnewcommand\INPUT{\item[{\textbf{Input:}}]}
\algnewcommand\OUTPUT{\item[{\textbf{Output:}}]}
\algrenewcommand{\algorithmiccomment}[1]{\hfill$\blacktriangleright$ #1}

\algsetblockdefx{MAPP}{ENDMAPP}{100}{0.5cm}[1]%
  {\INDSTATE \textbf{map partitions} #1}%
  {\INDSTATE \textbf{end map}}

\algsetblockdefx{MAP}{ENDMAP}{100}{0.5cm}[1]%
  {\INDSTATE \textbf{map} #1}%
  {\INDSTATE \textbf{end map}}

\algblockdefx{REDUCE}{ENDRED}[1]%
  {\textbf{reduce} #1}%
  {\textbf{end reduce}}

\begin{document}

\begin{frontmatter}

\title{BELIEF: A distance-based redundancy-proof feature selection method for Big Data.}



\author[ugr]{S. Ram\'irez-Gallego
\corref{mycorrespondingauthor}}
\ead{sramirez@decsai.ugr.es}
\cortext[mycorrespondingauthor]{Corresponding author}

\author[ugr]{S. Garc\'ia}
\ead{salvagl@decsai.ugr.es}
\author[mdh]{N. Xiong}
\ead{ning.xiong@mdh.se}
\author[ugr]{F. Herrera}
\ead{herrera@decsai.ugr.es}


\address[ugr]{Department of Computer Science and Artificial Intelligence, CITIC-UGR, University of Granada, 18071 Granada, Spain.}

\address[mdh]{School of Innovation, Design, and Engineering, M\"alardalen University V\"aster\aa s, SE-72123, Sweden.}

%\address[kau]{Faculty of Computing and Information Technology, King Abdulaziz University, Jeddah, Saudi Arabia.}

\begin{abstract}

With the advent of Big Data era, data reduction methods are highly demanded given its ability to simplify huge data, and ease complex learning processes. Concretely, algorithms that are able to filter relevant dimensions from a set of millions are of huge importance. Although effective, these techniques suffer from the ``scalability" curse as well. 

In this work, we propose a distributed feature weighting algorithm, which is able to rank millions of features in parallel using large samples. This method, inspired by the well-known RELIEF algorithm, introduces a novel redundancy elimination measure that provides similar schemes to those based on entropy at a much lower cost. It also allows smooth scale up when more instances are demanded in feature estimations. Empirical tests performed on our method show its estimation ability in manifold huge sets --both in number of features and instances--, as well as its simplified runtime cost (specially, at the redundancy detection step).

\end{abstract}


\end{frontmatter}

%\linenumbers

\section{Introduction}
\label{sec:intro}

% Big Data and the curse of dimensionality
Today the world keeps its relentless pace to the Big Data era by generating quintillion bytes of data daily. We have been surpassed by the challenge of processing such \textbf{volume} of data in a efficient and resilient way. Although researchers are devoting huge effort to cope with voluminous data from the \textit{instance} side, the opposite side has been largely disregarded by the community despite the more severe implications behind that. Zhai et. al deeply analyzed the dark side of volume in~\cite{zhai14}. In this study, authors smartly illustrate the current exponential grow of dimensions in public datasets (millions of features nowadays), as well as the lack of scalability of standard dimensionality reduction (DR) algorithms when facing real-world big data.

% lack of scalability, DR may be a fair solution
Recent developments in technology, science and industry have transformed the explosion of features into reality. Nevertheless, current machine learning (ML) libraries have lagged behind upward trends, thus demanding an urgent revision that enables a rapid learning from large-scale data. Although affected by the same pressing complexity~\cite{bolon15b}, DR techniques are seen as the most certain solution to the curse of dimensionality. They have been widely employed in smaller scenarios to simplify data wideness, and sometimes, have even served to increase subsequent learning performance~\cite{guyon06,zhao2011spectral,bolon15}.

% FS and FW
From the long taxonomy of DR techniques, the feature selection (FS) family can be highlighted by their high interpretability, popularity, and simplicity~\cite{qiu13,bolon14}. While FS algorithms requires an implicit parameter to delimit the magnitude of selection, feature weighting algorithms offer less constrained schemes based on feature-importance rankings. This subfamily of FS methods is specially relevant for large-scale contexts where interactive selection of thresholds is restricted. Additionally, some of the most outstanding models in FS are focused on feature weighting (e.g.: RELIEF~\cite{konon97}).

% problem of FS and blessing.
Some significant concerns must be addressed by feature selectors in Big Data, specially those based on pairwise correlations as they rapidly shift to trillions when facing millions of features. However, Zhai et al.~\cite{zhai14} also underpin the blessing side of the curse showing that in many problems most of features tend to contribute minimally with the correlation objective, and therefore, might be discarded.
%For all reasons cited above, this paper will mainly focus on FS algorithms. 

% scalable tools (Hadoop, Spark)
Several scalable tools and technologies have emerged in the last years to cope with Big Data. Most of them provide transparent high-level distributed processing services to end-users. MapReduce~\cite{dean04}, and its open-source version Hadoop~\cite{whi12,had15}, were the pioneer programming models in this area. Recently, a second generation of tools have come along aiming at amending the weaknesses of the first generation. For instance, disk-intensive specialization in Hadoop has being gradually shifted by Apache Spark~\cite{zaharia12,spa15}, a framework offering faster memory-based primitives. According to authors, Spark is aimed at accelerating interactive, online, and iterative procedures; normally present in ML algorithms. %This outstanding feature was really appreciated by data scientists given the iterative nature of learning algorithms.

% MLLIB, lack of FW methods
Big Data frameworks have given birth to a myriad of large-scale ML libraries, thus enabling standard ML algorithms to perform in huge databases so far unspoiled. MLlib~\cite{mll15}, for example, relies on Spark's operations to speed up the transition between iterations in ML. Although the current state of MLlib is quite advanced, it lacks from several data preprocessing~\cite{garcia16} algorithms. Focusing on dimensionality reduction, up to date only the $\chi^2$ selector, and an information-theoretic FS framework~\cite{ramirez17b} have been proposed. Still no real scalable feature weighting method has been integrated in Apache Spark up to now (see Section~\ref{subsec:others}).

In this work, we propose a distributed design for the feature weighting problem in huge scales inspired by the RELIEF-F algorithm. Our algorithm, called BELIEF, is specially optimized to address large-scale problems with millions of instances and features. Our main contributions with this work are as follows:

\begin{itemize}
	\item Bridging the gap of dimensionality reduction algorithms in large-scale ML libraries, such as MLlib. In this case, we provide an smart solution for the feature weighting subfield, up to now under-explored.
	\item Optimization of RELIEF's main algorithm with two novel procedures: one that squeezes the amount of data transferred during the neighborhood discovery step, and another that replaces instance-wise estimation by one based on the partition scope.
	\item Built-in redundancy removal to improve selection in problems dominated by feature redundancy. BELIEF takes advantage of feature distances to estimate inter-feature redundancy at nearly zero cost. Our solution based on co-occurrences offers similar schemes to those generated by state-of-the-art information-based measures. %Thus being one of the main drawbacks associated to original RELIEF-F addressed.
	\item Comprehensive empirical evaluation of BELIEF. This method has been compared to the current state-of-the-art in distributed FS, showing its advantage in terms of time and precision performance. Datasets with up to $O(10^7)$ instances and $O(10^4)$ features have been added to assess the theoretical scalability bounds of our proposal.

\end{itemize}
	
The main outline of the paper is as follows. First, some theoretical background information about FS, Big Data, and RELIEF have been captured in Section~\ref{sec:back}. Then Section~\ref{sec:method} describes how BELIEF works, as well as the main optimizations introduced to overcome scalability and redundancy problems. Detailed empirical results and a thorough analysis investigation are embedded in Section~\ref{sec:exp}. Finally, Section~\ref{sec:conc} outlines the concluding remarks derived from this work.

\section{Background}
\label{sec:back}

\subsection{Feature Selection: problem description and categorization}

In data preprocessing, FS algorithms~\cite{bolon15} focus on the task of isolating relevant and non-redundant features from the raw set of input features. The pursued objective here is to obtain a simpler and more sanitized subset of features that enables a proper generalization with minimum time cost and predictive degradation. Sometimes output models are not only simpler, but more accurate due to the associated nuances that usually disturb generalization are eliminated~\cite{guyon06}. 

Let $\mathbf{r_i}$ be an example $\mathbf{r_i} = (r_{i1}, \ldots, r_{in}, r_{iy})$ from $TR$ (the training set), where $r_{iy}$ value is associated with the output class $y \in C$, whereas $r_{i1}$ value corresponds to the $1$-th feature index, and the $i$-th sample index in the training dataset. $TR$ is formally defined as a bag of $m$ examples, whose instances $\mathbf{r_i}$ are formed by a set of input features $X$. Then, the FS problem is trivially defined as the task of finding a subset of features $S_\theta \subseteq X$ relevant for the mining process with minimum associated cost. 
%$S_\theta$ will then be a new scheme to train a new model in $D$ and to classify new patterns from $D_t$. 

The current FS state-of-the-art can be roughly summarized in three major categories according to~\cite{blum97}:

\begin{enumerate}
\item \textbf{Wrapper methods}: are FS algorithms that rely on a learning method to evaluate fitnesses of features~\cite{kohavi97}. Wrappers can be deemed as ad-hoc solutions, less prone to perform a proper generalization.
\item \textbf{Filtering methods}: are, on the other hand, independent methods that utilize external measures, such as separability or statistical dependences, to evaluate features. Their generality makes them more appropriate to learn directly from the explicit characteristics of datasets~\cite{guyon03}.
\item \textbf{Embedded methods}: are integrated methods that exploit a searching procedure implicit in the learning phase~\cite{saeys07}.
\end{enumerate}

Filtering usually imply a more general/better generalization due to their independence from learners. However, they usually are more conservative at removing features, and require more parameters. Some parameters are crucial for the learning task (e.g.: the number of selected features). They are also more efficient than wrappers, which is advantageous in Big Data environments. This fact explains why most of FS methods implemented in the Big Data literature are based on filtering techniques~\cite{garcia16}.

\subsubsection{Distance-based feature selection}

FS methods can alternatively be divided into two groups regarding the shape of output generated: either a subset of features, or a rated list of features (partial or complete)~\cite{bolon15}. The latter methods are called rankers, and are one of the most popular subcategory in filtering methods. Rankers rely on a given evaluation measure, such as one based on information dependency or distance, to measure and sort features by predictive importance. Evaluation is performed independently on each feature in these measures. However, although evaluation is performed feature-wisely, it is possible that individual scores somehow imply other features, for example, in redundancy-based measures. Efficient ranking methods are specially relevant for large-scale learning because of its efficiency and simplicity.

Distance measures used in FS range from Euclidean distance to more complex distances like Minkowski distance~\cite{garcia15}. Distance measures have not only been bounded to instance-based learning but also applied to class conditional density functions, such as Directed Divergence and Variance. One of the most popular FS solution based on distances is the RELIEF algorithm~\cite{kira92}, and its renovated version RELIEF-F~\cite{konon97}. Both are built upon feature distances computed to estimate relevance weights. 

RELIEFs are based on the idea that features are deemed relevant as long as they serve to distinguish close instances from alike classes. In RELIEF, we consider a feature $X_j \in X$ relevant if for a given instance $\mathbf{r_i}$ its near-hit (same class) neighbor $\mathbf{nh}$ is closer than its near-miss (distinct class) neighbor $\mathbf{nm}$ in the space defined by $X_j$. Feature-wise distances ($diff$) are accumulated in a weight vector $w$ of size $|X|$ as defined in Equation~\ref{eq:relief}. The parameter $s$ describes the size of the sampling set used in the estimation. The larger $s$, the more reliable is the approximation. This gives us a ranking where we tag as relevant those features whose associated weight exceeds an user-defined relevance threshold $\tau$:

\begin{equation}
\label{eq:relief}
w[X_j] = w[X_j] - diff(nh_{ij}, r_{ij}) / s + diff(nm_{ij}, r_{ij}) / s
\end{equation}

RELIEF-F~\cite{konon94} extended the original idea of miss-hit by expanding the neighborhood to all classes present in a given multi-class problem (Equation~\ref{eq:relieff}). Instead of relying on a single miss, RELIEF-F considers $k$ contributions for each opponent class $NM_{c}$. All contributions are weighted using their respective prior class likelihood as follows:
\begin{equation}
\label{eq:relieff}
w[X_j] = w[X_j] - \sum\limits_{i=1}^k \frac{diff(nh_{ij}, r_{ij})}{s \times k} + \sum_{c \neq y} \sum\limits_{i=1}^k \frac{[P(c) \times diff(NM_{cij}, r_{ij})]}{s \times k} 
\end{equation}

Beyond the multi-problem amendment, RELIEF was extended to deal with noise and missing data~\cite{konon94}. Complexity in all RELIEF versions is shared and determined by the neighbor search process, by definition $O(s \times m \times |X|)$. 

RELIEF is a reliable estimator whose effectiveness was proven by Kira et al. in~\cite{kira92}, among others~\cite{bolon2013review}. Authors showed that under some assumptions the expected weights for relevant features were much larger than those irrelevant. Nevertheless, RELIEF presents some drawbacks like the absence of an explicit mechanism for redundancy elimination. RELIEF directly obviates the final selection set might be fulfilled by both relevant and redundant features, such as those correlated or even duplicated (see Section~\ref{subsubsec:redundancy-relief}). 

\subsubsection{Redundancy elimination in RELIEF models}
\label{subsubsec:redundancy-relief}

%-Current techniques used to overcome redundancy elimination in RELIEF-F models.
%-General performance of these techniques and discussion about their possible implementation in distributed paradigms.

%SIMBA

%Relief does not re-evaluate the distances according to the weight vector w and thus it is inferior to Simba. In particular, Relief has no mechanism for eliminating redundant features. Simba may also choose correlated features, but only if this contributes to the overall performance.

In recent decades, several variations of RELIEF-F have been proposed. Most of them are focused on providing a solution to the redundancy problem, but there also exist others coping with noise. Here, we outline the most relevant contributions on the redundancy topic, as well as discuss about their possible adaptation to the Big Data environment.

In~\cite{bins01}, authors add a posterior phase based on K-means in order to filter out redundancy. K-means discovers cluster of correlated features, and selects those with highest scores as the pivotal elements. Correlation is then the distance measure elected for this algorithm.

Yang et al.~\cite{yang06} proposed the application of Gram-Schmidt orthogonal transformation to detect feature-pair correlations. The idea is to project one of the two feature vectors into the orthogonal dimension, and to re-compute relevance. Authors state that the cosine between the two original features highly influences the new relevance score, and therefore, it determines the correlation between both features. Based on cosine, the algorithm can estimate the redundancy relationship each pair of features.

Conditional RELIEF~\cite{wu12} (CRELIEF) amplifies the original distance-based score idea to one based on three interrelated measures: effectiveness (default), reliability, and informativeness. The latter factor addresses the redundancy problem by relating effectiveness with reliability of higher ranked features. Additionally, authors argue about the necessity of neighbors, and propose to substitute them by random tuples of instances. 

Wrapper solutions have also found its niche in the redundancy elimination topic. In~\cite{fu14}, Fu et. al proposes to polish RELIEF's output by applying Support Vector Machine Recursive Feature Elimination as secondary stage. The algorithm starts by dividing the original dataset into several groups in which a single instance of SVM is trained on each one. Feature scores are then normalized and eventually aggregated. 

Contrary to the solution above, EN-RELIEF~\cite{challi15} carries out an initial phase of redundancy removal through a regularized linear model. Elastic Net was chosen as the former solution because of its two-norm penalization (L1 and L2). After a post-RELIEF phase, EN-RELIEF generates a sparse vector reflecting the correlation between predictors.

Leaving aside wrappers, all filtering methods described above depends on one way or another on correlation to draw redundancy relationships between features. Notice that  the correlation sketch is generated by considering every pair interaction between features which gives us a total of $|X|^2$ iterations. In a Big Data context, this is unacceptable as the exponential growth of dimensions is a fact nowadays (see Section~\ref{subsec:bigdata}). Consequently, memory and time requirements of current redundancy elimination models should be softened either by reducing the amount of pairwise comparisons, or by alleviating their cost.

\subsection{Big Data overview: novel distributed processing tools and techniques and ``the curse of dimensionality"}
\label{subsec:bigdata}

%The ever-constant data generation in Internet is overwhelming the processing capabilities of existing systems in modern companies and public organizations. Additionally, the pressing need of extracting valuable information from such huge data has become a considerable challenge for data scientists and technical experts. 

Outstanding technologies, algorithms and tools are required to efficiently process what we call Big Data. The Big Data term was defined by Gartner~\cite{bey01} using the 3Vs concept, namely, high volume, velocity and variety information that require new processing schemes. The V's list was extended with 2 additional Vs (veracity and value) after a while.  

Gartner's scheme was intended to reflect the increasing number of examples in real-world problems, not envisaging the upcoming phenomenon of Big Dimensionality at least in the first epochs. This phenomenon, also called the ``Curse of Dimensionality''~\cite{zhai14}, revolves around the problem of exponential growth of features and its combinatorial impact in novel datasets. %The curse calls for new FS strategies and methods that cope with the feature explosion problem.

Little attention have been paid to the curse by ML community despite the exponential growth of dimensions is a fact in most of public data repositories (UCI or libSVM~\cite{bache13, chang11}) where it currently measures in the scale of millions. 

Despite explosion of correlations sharply affects FS algorithms, Zhai et al. proved in their study~\cite{zhai14} FS is in fact an affordable bless for large-scale processing. Studies on the News20 dataset show that features become more sparsely correlated as the dimensionality growths, which imply that the amount of correlation to be accounted is much lower than previously thought. Even so selecting relevant features from the raw set of potentially irrelevant, redundant and noisy features, while complying with the time and storage requirements, is one of the main challenges for scientists and practitioners nowadays. 

%MW From the machine learning's point view, this problematic has caused that many standard algorithms become obsolete. As result, new scalable methods capable of managing large-scale data are required. These algorithms do not have only to be scalable but also effective as the original methods that inspired them.

Big Data processing techniques and tools enable large-scale processing of data within tolerable elapsed times and precision ranges. Google was pioneer in this field by giving birth the MapReduce~\cite{dean04} framework in 2003. MapReduce automatically distributes the load among one or several machines in a easy and transparent way. Unlike other grid computing systems, MapReduce inherently addresses several technical nuances previously controlled by the user. The long list of duties assumed by MapReduce ranges from fault tolerance to network communication, among others.

The final user solely needs to implement two primitives (Map and Reduce) following a key-value scheme. During the Map stage, mappers threads read key-value pairs from local partitions, and transform them into a set of intermediate tuples eventually distributed according to a partitioning scheme (Equation~\ref{eq:map}). Normally coincident keys are sent to the same node. The Reduce phase read processed pairs and aggregated them to generate a summary result (Equation~\ref{eq:reduce}). For extra information about MapReduce and other distributed tools, or the design of distributed algorithms, please refer to~\cite{fern14} and~\cite{ramirez18}.

%The Map function takes $<$key, value$>$ pairs as input and yields a list of intermediate $<$key, value$>$ pairs as output. The Map function that internally processes the data is defined by the user following a key-value scheme. Hence, the general scheme for a Map function is defined as: 

\begin{equation}\label{eq:map}
Map(<key_1, value_1>) \rightarrow list(<key_2, value_2>)
\end{equation} 

%In the second phase, the master node groups pairs by key and distributes the combined result to the Reduce function launched in each node. Here, a reduction function is applied to each associated list value and a new output value is yielded. This process can be schematized as follows: 

\begin{equation}\label{eq:reduce}
Reduce(<key_2, list(value_2)>) \rightarrow <key_3, value_3>
\end{equation} 

Despite MapReduce's popularity in the ML field, the appropriateness of this framework for interactive or iterative processes has been highly criticized~\cite{lin13}. Disk-intensive processing in MapReduce makes it inappropriate or even inapplicable for most of current algorithms. 

Apache Spark~\cite{hamstra15,spa15} is a fast and general engine for large-scale data processing designed to overcome the drawbacks presented by Hadoop. Thanks to its built-in memory-based primitives, Spark is able to query data repeatedly deeming it suitable for iterative learning. According to the creators, Spark's engine is able to run up to 100 times faster than Hadoop in some cases.

Resilient Distributed Dataset (RDD) is the keystone structure on which Spark builds its workflow. RDD operators accompass and extend several distributed models like MapReduce by providing novel and more complex operators. They range from simple filtering and mapping processes to complex joins. As in MapReduce, Spark's primitives locally transform data within partitions, trying to maintain the data locality property as far as possible. RDDs also allows practitioners to customize data persistence, partitioning and data placement, broadcast read-only variables, track accumulators, and so on. For a full description of Spark operations, see~\cite{spa15}.

\section{BELIEF: an efficient distributed design for distance-based feature selection}
\label{sec:method}

In this section we present BELIEF, a distributed distance-based feature selection algorithm inspired by the popular RELIEF algorithm. %BELIEF's aim is to play an important role in the large-scale FS state-of-the-art as its predecessor did in the short-scale scenario.  
BELIEF has been implemented under the Apache Spark development framework to ensure that the two major iterative steps in RELIEF (neighbors searches and weight estimation) are optimized to their fullest degree in the distributed environment. 

Neighbor searches are solved in BELIEF by replicating the sample to all the nodes so that distances are completely computed in local (Section~\ref{subsec:nnsearch}). Neighborhood information for each instance is sent in form of locations to the partitions so that complete instances are not required to be sent. Secondly, BELIEF leverages the previous scheme to create a novel feature weighting estimation procedure where contributions are not instance-wisely anymore, but partition-wise. This mechanism extremely reduces the communication between partitions (Section~\ref{subsec:westimation}).

Besides the time performance enhancements, BELIEF also provides an efficient redundancy removal technique which leverages already computed feature distances to provide redundancy-based weights (Section~\ref{subsec:mcr}).

Algorithm~\ref{alg:main-belief} describes BELIEF's main procedure. It starts by dividing the sample set $S$ into several disjoint batches. Batches are employed in BELIEF for two reasons: to avoid the maximum size allowed by Spark's broadcasting be surpassed, and to create a feedback procedure between iterations which shorten the list of features to be considered in redundancy calculations. After the split phase, each batch feeds the relevance and redundancy functions. Partial relevance and redundancy matrices are then aggregated and passed to the Sequential Forward Selection (SFS) algorithm which will select features according to Equation~\ref{eq:both}.

\begin{algorithm}[!htb]
\small
\caption{BELIEF selection: Main procedure}
\label{alg:main-belief}
%\begin{multicols}{2}
\begin{algorithmic}[1]
\INPUT $D$ Dataset
\INPUT $s$ Sample size
\INPUT $b$ \# batches
\INPUT $k$ \# neighbors 
\INPUT $|S|$ \# features to be selected
\OUTPUT $S$ Set of selected features
\STATE $B \leftarrow$~sample($D, s$).split($b$)
\STATE $nfeat \leftarrow~|X|$
\STATE $O \leftarrow~vector(nfeat)$ \Comment{Marginal likelihood}
\STATE $P \leftarrow~matrix(nfeat, nfeat)$ \Comment{Joint likelihood}
\STATE $J \leftarrow~vector(nfeat)$ \Comment{Feature weights}
\FOR{$batch~\in~B$}
	 \STATE $query \leftarrow~$broadcast($batch$)
	 \STATE $NN \leftarrow~$broadcast(neighborhood($D, query, k$)) \Comment{Algorithm~\ref{alg:neighbors}}
	 \STATE $J_p, O_p, P_p \leftarrow~$weightEstimation($D, query, NN, k$) \Comment{Algorithm~\ref{alg:weights}}
     \STATE $J \leftarrow J + J_p;~O \leftarrow O + O_p;~P \leftarrow P + P_p$
\ENDFOR
\STATE $I_\alpha \leftarrow~$computeMCR($O,P$) \Comment{Equation~\ref{eq:mcr}}
\STATE $S \leftarrow$~SFS($J, I_\alpha, |S|$) \Comment{Equation~\ref{eq:both}}
\STATE return($S$)
\end{algorithmic}
%\end{multicols}
\end{algorithm}
 
\subsection{Nearest neighbor search in BELIEF}
\label{subsec:nnsearch}

As we mentioned before, the complexity order of RELIEF, and subsequently BELIEF, is mainly conditioned by the constant seek of neighbors ($\ell = O(s \cdot m)$). Once two examples are paired, weight computations are performed for each feature ($\ell^* = O(s \cdot m \cdot \mathbf{|X|})$).

Despite the widely recognized usefulness of RELIEF and other NN-based algorithms, neighbor searches are always compromised by each problem size because of two reasons: 

\begin{itemize}
	\item Execution time: for each search the entire dataset must be revisited. The task is not straightforwardly paralellizable (wide dependencies between partitions). The process becomes even more expensive in case sorting of $k > 1$ neighbors is required, which implies to maintain a dedicated structure, such as a bounded heap $O(k \cdot log(k))$.
	\item Memory consumption: As each pairwise computation must be taken into account in searches, the entire dataset is recommended to be allocated in heap memory. Otherwise, I/O disk operations will dominate global runtime.

\end{itemize}

The drawbacks mentioned above motivate novel designs for NN search which leverage the distributed technologies presented in Section~\ref{subsec:bigdata}, specially those based on in-memory operations.

In this work, we implement distributed searches following the scheme presented in kNN-IS~\cite{maillo17}. Assuming $TR$ and $TS$ are split and saved in $p$ disjoint partitions distributed across a cluster of $Z$ nodes. The MapReduce model divides the process into two stages: each mapper reads $p^* \leq p$ local partitions from $TR$ and the entire $TS$, computes the distances between each training partition and $TS$. Each reducer collects the local neighbors to each tuple partition-instance and aggregates them by selecting the closest neighbors to them. In our case, $TS$ is replaced by the sampling set in BELIEF.

kNN-IS has proven to be efficient in several real-world problems, however, it presents several bottlenecks to be analyzed. Firstly, the high communication cost derived from the replication of $TS$ to each node (memory consumption: $O(|TS| \cdot |X|$ per node); and secondly, the number of pairs sent to the reducers can become extremely large $p \cdot k \cdot |TS|$, although in this case it is not needed to send the complete instance but only the output variable (classification/prediction) and the distance value. In our case, complexity burden is even worse as $TS$ is replaced by the sampling set in BELIEF. Additionally, BELIEF requires the entire feature vector instead of only the output value.

The \textbf{reduce} step is then revisited in our proposal where entire instances are demanded. However, sending millions of input arrays across the network narrows as invalid, specially in high-dimensional scenarios. The solution adopted consists of sending a lightweight structure that defines the location of candidates in the partitioned dataset. The locator structure is defined as: an integer index pointing at its enclosing partition $I_{g}$, and another index $I_l$ for its local position within that partition. This trick reduces the memory and network consumption to $O(|TS| \cdot k \cdot p)$, where $|TS| = s$.

Once candidates are filtered, locators information is broadcasted to every node so that a single map phase can perform BELIEF's core estimations. Algorithm~\ref{alg:neighbors} lists the MapReduce process that describes this process. The following Section explains how the connection between neighbor searches and feature weight estimation.

\begin{algorithm}[!htb]
\small
\caption{Selection of nearest neighbors for the sample}
\label{alg:neighbors}
%\begin{multicols}{2}
\begin{algorithmic}[1]
\INPUT $D$ Dataset
\INPUT $S$ Sample set with size $s$ (broadcasted).
\INPUT $k$ \# number of neighbors 
\OUTPUT Neighbors locators $<index, list(<indexP, indexN>)>$
\MAPP{$<indexP, P>~\in~D$} 
	\FOR{$<index, input>~\in~S$}
		\FOR{$<indexN, inputN>~\in~P$}
			\STATE $d \leftarrow$ computeDistances($inputN$, $input$)
			\IF{isTopNN($d$)}
				\STATE addNN($index$)($indexP, indexN, d$)
			\ENDIF
		\ENDFOR
		\STATE emit($<index, addNN(index)>$) \Comment{addNN composed by: $<indexP, indexN, d>$}
	\ENDFOR
\ENDMAPP
\STATE \REDUCE{$<index, list(neighbors)> \in D$} 
\STATE $topNN \leftarrow$ selectTopNN(list($neighbors$))
\STATE emit($<index, topNN>$) \Comment{topNN composed by pairs: $<indexP, indexN>$}
\ENDRED
\STATE $return(SL)$
\end{algorithmic}
%\end{multicols}
\end{algorithm}
 
%~\cite{song15}

%Explain the model based on Maillo's implementation. In this case, the problem is tough because we need to retrieve all input features for each neighbor (use of "locators"). Use of batches, sampling and broadcasting.

\subsection{Global weights estimation in BELIEF}
\label{subsec:westimation}

The instance-wise estimation model integrated in RELIEF-F have proven to work well in small medical scenarios, such as tumor detection~\cite{konon94} or treatment of myopia~\cite{konon97}. However, its translation to big data scenarios is not straightforward because of the reasons exposed previously. Specifically, sending millions of arrays each time we need to update weights renders as extremely inefficient.

In this paper we propose to re-invent the original RELIEF formula, and to shift from an instance-wise estimation to a more scalable solution based on aggregating partial weights in a partition-wise manner. The pre-conditions we impose for the sake of scalability for this new formulation are: 

\begin{itemize}
	\item For each local process, each sampled instance will have only access to its local neighbors. No communication is allowed between processes. As the entire sample is replicated to all partitions, there will not be degradation in predictive performance.
	\item Instance-wise output contributions are not longer allowed. A compounding feature-wise solution for each data partition will be the new output.
\end{itemize}

In order to comply with the previous statements we define a new scheme for distance-based weight estimation, which is indeed applied in a single pass:

\begin{equation}
\label{eq:belief}
w[X_j] =  \sum_{i=1}^{|C|}{  \frac{DD_{ij}}{DC_{i}} \times P(C_i)} - \sum_{i=1}^{|C|}{ \frac{ED_{ij}}{EC_{i}} \times P(C_i) }
\end{equation}

\noindent where $DD$ and $ED$ are $|C| \times |X|$ matrices that summarize the accumulated feature distance between all sampled instances and its neighbors with \textit{distinct} and \textit{equal} class, respectively. $DC$ and $EC$ are $|C| \times 1$ matrices that counts the number of neighbors involved in the calculation of previous $DD$ and $ED$ matrices, respectively.

Matrix computation and weight estimation are computed through two different MapReduce processes described in Algorithm~\ref{alg:weights}. The first phase relies on Equation~\ref{alg:neighbors} to compute neighbors' locations. Then, it creates the feature-class matrices which are updated with distance information. Matrices are eventually aggregated at the subsequent reduce phase. Finally, another MapReduce process (with no reducers) is programmed to apply Equation~\ref{eq:belief} to each set of matrices. 

\begin{algorithm}[!htb]
\small
\caption{RELIEF's feature weight estimation}
\label{alg:weights}
%\begin{multicols}{2}
\begin{algorithmic}[1]
\INPUT $D$ Dataset 
\INPUT $S$ Sample set with size $s$ (broadcasted).
\INPUT $NL$ Neighbor locators (broadcasted)
\OUTPUT Weight by feature 
\MAPP{$<indexP, P>~\in~D$} 
	\STATE $DD \leftarrow $matrix($|C|,|X|$); $ED \leftarrow $matrix($|C|,|X|$)
	\STATE $DD \leftarrow $matrix($|C|, 1$); $EC \leftarrow $matrix($|C|, 1$)
	\FOR{$<index, input, label>~\in~S$}	
		\STATE $indices \leftarrow NL$.getLocalLocators($index, indexP$)
		\FOR{$i~\in~indices$}
			\FOR{$j~\in~|X|$}
				\STATE $distance \leftarrow$ diff($P(i)$.getInput($j$), $input(i)(j)$)
				\IF{$P(i)$.label $\neq$ $label$}
					\STATE $DD(label)(j) \leftarrow distance$ 
				\ELSE
					\STATE $ED(label)(j) \leftarrow distance$ 			
				\ENDIF 
			\ENDFOR
			\IF{$P(i)$.label $\neq$ $label$}
				\STATE $DC(label) \leftarrow DC(label) + 1$ 
			\ELSE
				\STATE $EC(label) \leftarrow EC(label) + 1$ 			
			\ENDIF 
		\ENDFOR
	\ENDFOR
	\FOR{$j~\in~|X|$}
		\STATE $matrices \leftarrow <DD(*)(j),ED(*)(j),DC(j),EC(j)>$
		\STATE emit($<j, matrices>$)
	\ENDFOR
\ENDMAPP
\STATE \REDUCE{$<feature, list(matrices)>$} 
\STATE sumMatrices(list($matrices$))
\ENDRED
\MAP{$<feature, matrices>~$} 
\STATE $weight \leftarrow $ applyBELIEF($matrices$)
\STATE emit($<feature, weight>$)
\ENDMAP
\end{algorithmic}
%\end{multicols}
\end{algorithm}

BELIEF's approach gains scalability and efficiency power with respect to RELIEF, while precision performance remains similar. Although in BELIEF the estimation scope is extended beyond individual instances, our solution keeps unaltered the main idea behind RELIEF. Conceptually the BELIEF method differs from its predecessor in several aspects: 

\begin{enumerate}
	\item In RELIEF each matrix cell would involve an exact $k$ number of neighbors, whereas in BELIEF we resort to standard k-NN search to avoid searches in too broad areas. By weighting each cell value by its neighborhood size this problem is partially addressed.
	\item Class likelihood weighting has been extended in BELIEF to the negative part (right side of Equation~\ref{eq:belief}). We understand this model is much more natural and separable for further computations than that presented by RELIEF.
	\item As mentioned before, the main improvement introduces revolves around the use of single-pass estimation based on individual feature-class contributions. Though there is substantial shift between both models, BELIEF mimics the same idea based on class separability held in RELIEF. Also notice that some repetitive factors in Equation~\ref{eq:relief} can be easily removed since they appear in each instance-wise sum, for example, class likelihood or the sample size $s$. In fact, $s$ provides nothing relevant beyond a simple normalization.
\end{enumerate}

%Explain the design that transforms the instance-based evaluation model to
%a global-based model based on partial solution summarization.
%
%Formulation for BELIEF.
%
%
%Introducción con la fórmula de relief. Hablamos de que el método de RELIEF por instancia no es compatible con el método previo de cálculo del vecindario. Es necesario una nueva fórmula que posibilite el cálculo de pesos de una manera más global. Proponemos la fórmula de BELIEF. Pseudocódigo.

\subsection{mCR (minimum Collision-based Redundancy): an efficient redundancy removal technique for BELIEF}
\label{subsec:mcr}

In Section~\ref{subsubsec:redundancy-relief}, we have enumerated different techniques that enable redundancy elimination in RELIEF-F algorithms. Nevertheless, as stated in this section, all these techniques have been designed for small scenarios.

A possible scalable solution for redundancy control may be one based on information theory. An uncertainty measure widely used in the literature is Mutual Information (MI)~\cite{cover91}, which expresses the loss of uncertainty of one variable $X_i$ after knowing other random variables. MI can be rewritten in entropy terms as follows:

\begin{equation}
\label{eq:mi}
\begin{aligned}
I(X_i;X_j) &= H(X_i) - H(X_i|X_j) \\
       &= \sum_{a\in X_i} \sum_{b\in X_j} P(a,b) \log \frac{P(a,b)}{P(a) P(b)}.
\end{aligned}
\end{equation}

\noindent where $X_i$ and $X_j$ are two discrete random variables with marginal probability mass functions $P(a)$ and  $P(b)$, respectively. $H$ represents Shannon entropy, and $P(a,b)$ a joint mass function.

In FS, those features that bears similar information according to MI are considered as redundant, and consequently can be discarded. Some relevant FS filters, like minimum Redundancy Maximum Relevance~\cite{peng2005}, rely on these information-based measures to make a trade-off with relevancy. 

The main drawback of information theoretical techniques is their high complexity. All available combinations between each pair of features (joint likelihood), and all single occurrences in each single feature (marginal likelihood) are accounted for weight estimation, which supposes an unbearable cost in some large-scale scenarios~\cite{ramirez17b}.

%En la sección \ref{subsubsec:redundancy-relief} hemos presentado diversas técnicas which make RELIEF able to remove redundancy from its schemes. Su principal problema es que fueron diseñadas paara problemas pequeños y no fit en el escenario de big data.

%Una posible solucion sería una basada en medidas de redundancia basadas en información como las usadas pro técnicas como MRMR, el problema principal es que el cálculo de probabilidades implica considera todas las combinaciones entre atributos (para las joint) y aquellas dentro del atributo (marginal). Todo esto para cada pair de atributos. La formula usada para la redundancia en mrmr is la conditional entropia :

From the previous formulation we can deduce that entropy is mainly dominated by those co-occurrences more recurrent in the series. Influence of isolated values is then almost negligible. Furthermore, since concrete values in co-occurrences are no longer accounted after being subsumed by the formula, we suggest replacing standard MI by a measure based on directly measuring the number of ``collisions" or co-occurrences. This though may imply loss of information and proficiency, it will surely simplifies matrices and computations. After applying Shannon's entropy to the new ``collision" variable, we obtain $I_\alpha$:

%Random variable $X_i$ in MI is then substituted by the collision variable.

%está eminentemente marcado por aquellas combinaciones que se repiten más en la serie. Además dado que la formula ignora el origen de las probabilidades una vez que éstas han sido agregados, se podría pensar en incluir una medida basada en colisiones en vez de combinaciones. La idea es sustituir la probabilidad de una variable aleatoria X por la probabilidad de colisión. De esta manera reemplazamos la sumatoria de combinaciones individuales por el número total de combinaciones coincidentes (colisiones). Tras aplicar la definición de Shanon obtendríamos:

%Entropy is eminently mar

\begin{equation}
\label{eq:mcr}
\begin{aligned}
I_\alpha(X_i;X_j) = PC(X_i) \log \frac{PC(X_i, X_j)}{PC(X_i) PC(X_j)}.
\end{aligned}
\end{equation}

\noindent where $PC$ represents the likelihood of coincidence within any pair of input feature and/or a single one. We call this measure \textbf{minimum Collision-based Redundancy (mCR)}.

%Description of the technique based on collisions. General formulation. Integration with distance weights via maxmin-norm.

%Discussion about how the method included in BELIEF, which is based on collisions, may offer similar results to other InfoTheoretical-based methods, but with much better performance as it does not consider all combinations.

mCR can be easily integrated with BELIEF by normalizing both measures and summing their contributions. In our experiments we rely on minmax normalization and a weighting factor $\theta$ to relate both factors. mCR is designed to be integrated in a Sequential Forward Selection process~\cite{garcia15} where we start with an empty set of features $S$, and select the best feature in each epoch according to a criteria $J$ until $|S|$ features are selected. Ranks for non-selected features are updated in each iteration taking as reference the last feature selected, and following the formula:

\begin{equation}
\label{eq:both}
J(X_i) = w(X_i) - \theta \sum_{X_j \in S} I_\alpha(X_j;X_i),
\end{equation}

\noindent where $\theta$ is the factor that weights the impact of redundancy (mCR) and relevance (BELIEF).

One of the most relevant advantages in mCR is that it leverages prior distance values computed in the previous step to construct $PC$ matrices. No extra cost is then associated to mCR beyond the annotation of joint coincidences. Although the number of annotations is infrequent, if this fact is left unmanaged mCR will endure the same problems presented by information-based measures (Section~\ref{subsubsec:redundancy-relief}). 

In order to control the magnitude of accounted collisions, we introduce a new parameter that limits the number of features considered. The idea is to only update $PC(X_i, X_j)$ iff one of them is ranked in top-$(|S| \cdot \eta)$ features by the previous BELIEF phase. It makes sense to leave high irrelevant features aside as they will surely not overtake relevant features in the ranking after the redundancy update. We propose a default value of 2.0 for $\eta$.

mCR is thought to be applied for discrete features where collisions can be easily accounted. However, most of real-world problems partially or entirely consists of continuous features. For continuous scenarios, we propose an alternative solution that replaces 0,1 updates (1 collision hit, 0 otherwise) by a percentage that measures the magnitude of collision, called collision rate $CR$ and defined as follows:

\begin{equation}
\label{eq:criteria}
CR = 1 - [(r_{1i} - r_{2i}) / \uparrow CR_{X_i}]
\end{equation}

\noindent where $r_1$ and $r_2$ are two neighbors selected by BELIEF, 
$\uparrow CR_{X_i}$ the maximum collision rate for $X_i$ with $\uparrow CR = 6\sigma_i$ for all the input features, and $\sigma_i$ the standard deviation for $X_i$. 

This decision is motivated by the Chebyshev's inequality rule which states that 89\% of values in most probability distributions are within three standard deviations of the mean. Given that we only focus on the higher collision rates, Chebyshev's inequality let us to safely ignore outliers, namely, those with the lowest collision values. Another possible solution is to define $\uparrow CR$ equal to the maximum range for each feature, however, this option is highly affected by the shape of distributions.

In order to reduce some effort on annotating coincidences, we establish an upper limit $\kappa$ for collision rates so that values below $\kappa$ are directly skipped. %Given that we only focus on close-to-zero distances, we can state that values out of range are . 
Accepted rates are then utilized to update $PC$ matrices following the scheme $[0 \cup [\kappa,1]]$. On the other hand, we apply a Z-score normalization to simplify $\uparrow CR = 6\sigma = 6$ which improves the homogeneity between features, and at the same time the performance of neighbor searches. 

\subsection{Other RELIEF adaptations for Big Data tools}
\label{subsec:others}

In~\cite{palma18}, authors proposed a distributed ReliefF-based solution for Apache Spark, called DiReliefF. Despite the algorithm has been successfully tested on real large-scale datasets, authors made some assumptions about the estimation sample which can be deemed as unfair. Namely, they assert that tiny samples with few hundreds of instances are enough to properly estimate class separability in problems formed by millions of instances. As an example, authors states that $6.25 \times 10^-7\%$ of data in the ECBDL14 problem (see Section~\ref{subsec:framework}) is enough to correctly underpin feature weights. From our point of view, this premise seems unrealistic as the chance of properly representing millions of instances with such small sample renders as negligible. Experiments focusing on proving the reliability of previous premise will be performed in Section~\ref{subsec:sampling-impact}.

Previous assumptions served as a basement for the optimizations introduced in DiReliefF. For instance, neighborhoods in DiReliefF are computed and moved across the network in their original shape. Although intuitive this procedure impose a high communication cost whenever the number of features or estimation samples increases. Similar cost is imposed by the last step in DiReliefF when neighborhoods are pushed to the driver node to perform feature-side averages. Again this process is simple, but hardly scalable and resource-wasting. 

For the above reasons, we think a novel distributed design of ReliefF based on realistic network optimizations, and further enriched with redundancy control techniques can be of great interest for the literature. 

\section{Empirical evaluation}
\label{sec:exp}

\subsection{Experimental framework}
\label{subsec:framework}
BELIEF has been tested on four large-scale classification datasets, grouped in two categories according to their format and shape. \textit{ECBDL14} and \textit{epsilon} are dense datasets with tabular format, a large number of examples, and a medium number of features. \textit{url} and \textit{kddb} are two sparse dataset (from the libSVM repository) formed by millions of key-value pairs. 

\textit{ECBDL14} is a binary imbalance dataset with an oversampled training set of 65 millions of instances and 631 input features. It is specially remarkable the relevance and difficulty of~\textit{ECBDL14} given the high imbalance ratio present in the original training set (98\%). The remaining datasets are hosted in the LibSVM dataset repository~\cite{chang11}. Their origin, format, and other information can be found in the project's website.\footnote{\url{http://www.csie.ntu.edu.tw/~cjlin/libsvmtools/datasets/}}. Table~\ref{tab:datasets} provides basic information about the size and magnitude of all problems. 

\begin{table}[!htp]
\renewcommand{\arraystretch}{1.3}
\centering
\scriptsize
\caption{Basic information about the datasets. Included: number of examples for training and test sets (\#Train Ex., \#Test Ex.), number of features (\#Atts.), number of output labels (\#Cl) and sparsity condition (binary).}
\label{tab:datasets}
\resizebox{0.9\textwidth}{!}{
  \begin{tabular}{|l||c||c||c||c||c|}
  \hline {\bf Data Set} & {\bf \#Train Ex.} & {\bf \#Test Ex.} & {\bf \#Atts.}  &{\bf \#Cl.} & {\bf Sparse}\\
  \hline
epsilon & 400 000 & 100 000 & 2000 & 2 & No \\
%dna & 79 739 293 & 10 000 000 & 200 & 2 & No\\
ECBDL14 & 65 003 913 & 2 897 917 & 630 & 2 & No\\
url & 1 916 904 & 479 226 & 3 231 961 & 2 & Yes\\
kddb & 19 264 097 & 748 401 & 29 890 095 & 2 & Yes\\
  \hline
  \end{tabular}
}
\end{table}

%% TODO

For comparison purposes we have included a distributed exact version of the minimum Redundancy Maximum Relevance algorithm (DmRMR)~\cite{peng2005,ramirez17b}, implemented in Apache Spark. DmRMR inclusion aims at showing pros and cons of both alternatives, as well as demonstrate the validity of BELIEF. The same argument was proven in~\cite{bolon2013review} where the standard version of both algorithms were compared in a large list of small synthetic datasets.

FS schemes are evaluated using two classification algorithms belonging to the MLlib library~\cite{mll15}: Support Vector Machines (SVM) and Decision Trees (DT). SVMs in Spark internally optimizes the Hinge Loss using Orthant-Wise Limited-memory Quasi-Newton optimizer, whereas DTs perform recursive binary partitioning optimizing an information gain measure (Gini impurity or InfoGain). Parameter configuration for BELIEF is shown in Table~\ref{tab:parameters}. Default values for classifiers are left alone. Since all selectors and predictors are based on iterative processes, we cached all the training sets in memory at the beginning.

\begin{table}[!htp]
\renewcommand{\arraystretch}{1.3}
\centering
\scriptsize
\caption{Parameters configuration for selectors}
\label{tab:parameters}
\resizebox{0.75\textwidth}{!}{
  \begin{tabular}{|l|c|}
  \hline
  \textbf{Method} & \textbf{Parameters}\\
  \hline
  BELIEF & k = 3, 5, 10 \\
  BELIEF & sampling rate (s) = 0.01, 0.02, 0.25, 0.5 \\
  BELIEF & batch size (bs) = 0.1, 0.25 \\
  BELIEF & $|S|$ = 10, 50, 100 \\ 
  DmRMR	& $|S|$ = 10, 50, 100 \\ 
  BELIEF \& DmRMR & Spark partitions = 920\\
  \hline
  \end{tabular}
  }
\end{table}

%% TODO

The default level of parallelism was established to 2 times (920) the number of virtual threads available in the cluster (460), thus following the guidelines stated by Spark's creators~\footnote{\url{http://spark.apache.org/docs/latest/programming-guide.html#parallelized-collections}}.
%For the comparison in Section~\ref{subsec:results-time}, this value will be changed (see details below).

F1 score (harmonic mean of precision and recall) and prediction accuracy are the two evaluation metrics elected to assess the utility of the selection schemes. To evaluate time performance we rely directly on the overall cluster prediction and feature selection time in seconds.

The cluster involved in the large-scale experiments is composed by 20 slave nodes and 1 master node. The computing nodes hold the following features: 2 CPU processors x Intel Xeon E5-2620, 6 real cores per CPU, 2.00 GHz, 15 MB cache, QDR InfiniBand Network (40 Gbps), 2 TB HDD, 64 GB RAM. All of them running the following software: Apache Spark and MLlib 2.2.0, Hadoop 2.6.0 (HDFS replication factor 2, HDFS default block size 128 MB), 460 virtual threads, 960 RAM GB.

%Both the HDFS and Spark master processes (the HDFS NameNode and the Spark Master) were hosted in the main node. The NameNode controlled the HDFS and coordinated the slave machines by means of their respective DataNode daemons. The Spark Master controlled all the executors in each worker node. Spark used the HDFS file system to load and save data in the same way as the Hadoop framework. 

For reproducibility purposes the code have been opensourced and uploaded to GitHub~\url{https://github.com/sramirez/spark-RELIEFFC-fselection}. In the close future we will send a request for its integration in the main Spark API.

\subsection{BELIEF (and mCR) evaluation on small controlled environments}

Analysis starts with the evaluation of BELIEF and mCR in small and synthetic problems where relevancy and redundancy is known and well-defined, thus being easier to study algorithms' behaviors. The same study proposed in a FS review~\cite{bolon2013review} is replicated here in a smaller scale. Table~\ref{tab:small-datasets} shows the main characteristics of the synthetic datasets used~\cite{bolon2013review}, as well as, the composition and nature (relevant/redundant/noise) of synthetic features.

\begin{table}[!htp]
\renewcommand{\arraystretch}{1.3}
\centering
\scriptsize
\caption{Basic information about the synthetic datasets from~\cite{bolon2013review}. From left to right: dataset name, number of rows, number of columns, list of relevant and redundant features, and the baseline accuracy obtained with no FS. (*) SD3 presents six groups of 10 features created to be redundant among themselves; the other 4,000 features are irrelevant. The ideal output is that with one feature from each group.}
\label{tab:small-datasets}
\resizebox{0.99\textwidth}{!}{
\begin{tabular}{l|c|c|c|c|c}
\hline
\textbf{Dataset} & \multicolumn{1}{l|}{\textbf{\# features}} & \multicolumn{1}{l|}{\textbf{\# samples}} & \textbf{Relevant} & \textbf{Redundant} & \multicolumn{1}{l}{\textbf{Baseline acc.}} \\ \hline
Corral-100 & 99 & 32 & 1-4 & -- & 56.25\% \\
XOR-100 & 99 & 50 & 1,2 & -- & 52.00\% \\
Parity-3+3 & 12 & 64 & 1-3 & 4-6 & 50.00\% \\
%SD1 & 4020 & 75 & G1,G2 & others & 33.33\% \\ 
%SD2 & 4040 & 75 & G1-G4 & others & 33.33\% \\
SD3 & 4060 & 75 & G1-G6 & * & 33.33\% \\ 
Madelon & 500 & 2400 & 1-5 & 6-20 & 50.13\% \\ \hline
\end{tabular}
}
\end{table}

Besides accuracy evaluation, we furthermore analyze the composition of the FS schemes generated by using a scoring measure proposed in~\cite{bolon2013review}:

\begin{equation}
Suc. = [\frac{S_{rel}}{X_{rel}} - \zeta \frac{S_{red}}{X_{red}}] \times 100
\end{equation}

\noindent where $S_{rel}, X_{rel}$ is the number of relevant features in $S$ and $X$, respectively. The remaining variables stand for redundant features. The $Suc.$ score described above was designed to penalize redundant features, and to reward relevant selections.

Tables~\ref{tab:corral} --~\ref{tab:madelon} contains diverse performance information concerning composition, accuracy and success obtained by schemes in mRMR, BELIEF and BELIEF + mCR in five datasets. Specially relevant is the column \textbf{\# red} which indicates the degree of redundancy cleaning achieved by each measure. Results shed some light about the potential of the conjunction mCR and BELIEF. This combination overcomes its competitors in 3/5 datasets, being specially relevant in 2/3 datasets with redundant features. 

\begin{table}[!htp]
\renewcommand{\arraystretch}{1.3}
\centering
\scriptsize
\caption{Evaluation results for Corral-100. From left to right: the method name, the total number of features selected, and which and how many relevant, redundant and irrelevant are selected. The right-most part contains the prediction results using Na\"ive Bayes, Decision Tree, and Logistic Regression. Highlighted results in bold indicate the best outcome for each FS scheme.}
\label{tab:corral}
\resizebox{0.99\textwidth}{!}{
\begin{tabular}{c|r|c|c|c|c|r|r|r|r}
\hline
\multicolumn{1}{l|}{\textbf{}} & \multicolumn{ 6}{c|}{\textbf{Feature Selection}} & \multicolumn{ 3}{c}{\textbf{Accuracy}} \\ \hline
\multicolumn{1}{c|}{\textbf{Method}} & \multicolumn{1}{l|}{\textbf{\# sel.}} & \multicolumn{1}{l|}{\textbf{Rel.}} & \multicolumn{1}{l|}{\textbf{\# rel.}} & \multicolumn{1}{l|}{\textbf{\# red.}} & \multicolumn{1}{l|}{\textbf{\# irrel.}} & \multicolumn{1}{l|}{\textbf{Success}} & \multicolumn{1}{l|}{\textbf{NB}} & \multicolumn{1}{l|}{\textbf{DT}} & \multicolumn{1}{l}{\textbf{LR}} \\ \hline
\multicolumn{ 1}{c|}{\textbf{BELIEF}} & \textbf{\textit{4}} & 1,3 & 2 & - & 2 & \textbf{0.50} & 0.7105 & 0.6605 & 0.6716 \\ 
\multicolumn{ 1}{c|}{} & \textbf{\textit{10}} & 1-3 & 3 & - & 7 & 0.75 & 0.8077 & 0.6405 & 0.6216 \\ \hline
\multicolumn{ 1}{c|}{\textbf{BELIF+mCR}} & \textbf{\textit{4}} & 3 & 1 & - & 3 & 0.25 & 0.6155 & 0.7266 & 0.7627 \\ 
\multicolumn{ 1}{c|}{} & \textbf{\textit{10}} & 1-3 & 3 & - & 7 & 0.75 & 0.7466 & 0.7516 & 0.7066 \\ \hline
\multicolumn{ 1}{c|}{\textbf{mRMR}} & \textbf{\textit{4}} & 3 & 1 & - & 3 & 0.25 & 0.6072 & 0.7266 & \textbf{\textit{0.7716}} \\ 
\multicolumn{ 1}{c|}{} & \textbf{\textit{10}} & 1-4 & 4 & - & 6 & \textbf{1.00} & 0.7655 & 0.6316 & \textbf{\textit{0.8438}} \\ \hline
\end{tabular}
}
\end{table}

\begin{table}[!htp]
\renewcommand{\arraystretch}{1.3}
\centering
\scriptsize
\caption{Evaluation results for XOR-100. From left to right: the method name, the total number of features selected, and which and how many relevant, redundant and irrelevant are selected. The right-most part contains the prediction results using Na\"ive Bayes, Decision Tree, and Logistic Regression. Highlighted results in bold indicate the best outcome for each FS scheme.}
\label{tab:xor}
\resizebox{0.99\textwidth}{!}{
\begin{tabular}{c|r|c|c|c|c|r|r|r|r}
\hline
\multicolumn{1}{l|}{\textbf{}} & \multicolumn{ 6}{c|}{\textbf{Feature Selection}} & \multicolumn{ 3}{c}{\textbf{Accuracy}} \\ \hline
\multicolumn{1}{l|}{\textbf{Method}} & \multicolumn{1}{l|}{\textbf{\# sel.}} & \multicolumn{1}{l|}{\textbf{Rel.}} & \multicolumn{1}{l|}{\textbf{\# rel.}} & \multicolumn{1}{l|}{\textbf{\# red.}} & \multicolumn{1}{l|}{\textbf{\# irrel.}} & \multicolumn{1}{l|}{\textbf{success}} & \multicolumn{1}{l|}{\textbf{NB}} & \multicolumn{1}{l|}{\textbf{DT}} & \multicolumn{1}{l}{\textbf{LR}} \\ \hline
\multicolumn{ 1}{c|}{\textbf{BELIF}} & \textbf{\textit{2}} & - & 0 & - & 2 & 0.00 & 0.7440 & \textbf{\textit{1.0000}} & 0.6204 \\ 
\multicolumn{ 1}{c|}{} & \textbf{\textit{10}} & - & 0 & - & 10 & 0.00 & 0.7197 & 0.8383 & 0.7840 \\ \hline
\multicolumn{ 1}{c|}{\textbf{BELIF+mCR}} & \textbf{\textit{2}} & - & 0 & - & 2 & 0.00 & 0.7440 & \textbf{\textit{1.0000}} & 0.6204 \\ 
\multicolumn{ 1}{c|}{} & \textbf{\textit{10}} & - & 0 & - & 10 & 0.00 & 0.7190 & \textbf{\textit{0.9040}} & 0.6840 \\ \hline
\multicolumn{ 1}{c|}{\textbf{mRMR}} & \textbf{\textit{2}} & - & 0 & - & 2 & 0.00 & 0.4854 & 0.7157 & 0.6957 \\
\multicolumn{ 1}{c|}{} & \textbf{\textit{10}} & - & 0 & - & 10 & 0.00 & 0.7116 & 0.6173 & 0.6866 \\ \hline
\end{tabular}
}
\end{table}

\begin{table}[!htp]
\renewcommand{\arraystretch}{1.3}
\centering
\scriptsize
\caption{Evaluation results for Parity-3+3. From left to right: the method name, the total number of features selected, and which and how many relevant, redundant and irrelevant are selected. The right-most part contains the prediction results using Na\"ive Bayes, Decision Tree, and Logistic Regression. Highlighted results in bold indicate the best outcome for each FS scheme.}
\label{tab:parity}
\resizebox{0.99\textwidth}{!}{
\begin{tabular}{c|r|c|c|c|c|r|r|r|r}
\hline
\multicolumn{1}{l|}{\textbf{}} & \multicolumn{ 6}{c|}{\textbf{Feature Selection}} & \multicolumn{ 3}{c}{\textbf{Accuracy}} \\ \hline
\multicolumn{1}{l|}{\textbf{Method}} & \multicolumn{1}{l|}{\textbf{\# sel.}} & \multicolumn{1}{l|}{\textbf{Rel.}} & \multicolumn{1}{l|}{\textbf{\# rel.}} & \multicolumn{1}{l|}{\textbf{\# red.}} & \multicolumn{1}{l|}{\textbf{\# irrel.}} & \multicolumn{1}{l|}{\textbf{success}} & \multicolumn{1}{l|}{\textbf{NB}} & \multicolumn{1}{l|}{\textbf{DT}} & \multicolumn{1}{l}{\textbf{LR}} \\ \hline
\multicolumn{ 1}{c|}{\textbf{BELIEF}} & \textbf{\textit{3}} & 1,2 & 2 & 1 & 0 & 0.63 & 0.3045 & 0.2902 & 0.2902 \\ 
\multicolumn{ 1}{c|}{} & \textbf{\textit{5}} & 1-3 & 3 & 2 & 0 & \textbf{0.93} & 0.3045 & \textbf{\textit{0.8914}} & 0.2645 \\ \hline
\multicolumn{ 1}{c|}{\textbf{BELIEF+mCR}} & \textbf{\textit{3}} & 1-3 & 3 & 0 & 0 & \textbf{1.00} & 0.3245 & \textbf{\textit{0.8914}} & 0.2645 \\ 
\multicolumn{ 1}{c|}{} & \textbf{\textit{5}} & 1-3 & 3 & 2 & 0 & \textbf{0.93} & 0.3045 & \textbf{\textit{0.8914}} & 0.2645 \\ \hline
\multicolumn{ 1}{c|}{\textbf{mRMR}} & \textbf{\textit{3}} & - & 0 & 0 & 3 & -0.11 & 0.6195 & 0.5160 & 0.6170 \\ 
\multicolumn{ 1}{c|}{} & \textbf{\textit{5}} & 2,3 & 2 & 0 & 3 & 0.56 & 0.5795 & 0.4704 & 0.6295 \\ \hline
\end{tabular}
}
\end{table}

\begin{table}[!htp]
\renewcommand{\arraystretch}{1.3}
\centering
\scriptsize
\caption{Evaluation results for SD3. From left to right: the method name, the total number of features selected, and which and how many relevant, redundant and irrelevant are selected. The right-most part contains the prediction results using Na\"ive Bayes, Decision Tree, and Logistic Regression. Highlighted results in bold indicate the best outcome for each FS scheme. Rows with asterisk mean that a previous discretization phase is performed to enable NB prediction.}
\label{tab:sd3}
\resizebox{0.99\textwidth}{!}{
\begin{tabular}{c|r|c|c|c|r|r|r|r}
\hline
\multicolumn{1}{l|}{\textbf{}} & \multicolumn{ 5}{c|}{\textbf{Feature Selection}} & \multicolumn{ 3}{c}{\textbf{Accuracy}} \\ \hline
\multicolumn{1}{l|}{\textbf{Method}} & \multicolumn{1}{l|}{\textbf{\# sel.}} & \multicolumn{1}{l|}{\textbf{\# rel.}} & \multicolumn{1}{l|}{\textbf{\# red.}} & \multicolumn{1}{l|}{\textbf{\# irrel.}} & \multicolumn{1}{l|}{\textbf{success}} & \multicolumn{1}{l|}{\textbf{DT}} & \multicolumn{1}{l|}{\textbf{LR}} & \multicolumn{1}{l}{\textbf{NB}} \\ \hline
\multicolumn{ 1}{c|}{\textbf{BELIEF}} & \textbf{\textit{6}} & 1 & 5 & 0 & 0.17 & 0.5205 & 0.5171 & \multicolumn{1}{c}{-} \\ 
\multicolumn{ 1}{c|}{} & \textbf{\textit{20}} & 2 & 8 & 10 & 0.33 & 0.4927 & 0.6344 & \multicolumn{1}{c}{-} \\ \hline
\multicolumn{ 1}{c|}{\textbf{BELIEF+mCR}} & \textbf{\textit{6}} & 1 & 1 & 4 & 0.17 & 0.5447 & 0.6033 & \multicolumn{1}{c}{-} \\ 
\multicolumn{ 1}{c|}{} & \textbf{\textit{20}} & 2 & 7 & 11 & 0.33 & 0.4821 & 0.7053 & \multicolumn{1}{c}{-} \\ \hline
\multicolumn{ 1}{c|}{\textbf{mRMR}} & \textbf{\textit{6}} & 2 & 0 & 4 & \textbf{0.33} & \textbf{\textit{0.8815}} & 0.8758 & 0.7831 \\ 
\multicolumn{ 1}{c|}{} & \textbf{\textit{20}} & 4 & 3 & 13 & \textbf{0.67} & 0.6562 & \textbf{\textit{0.9423}} & 0.7921 \\ \hline
\multicolumn{ 1}{c|}{\textbf{BELIEF *}} & \textbf{\textit{6}} & 2 & 4 & 0 & \textbf{0.33} & 0.5780 & 0.5080 & 0.1225 \\
\multicolumn{ 1}{c|}{} & \textbf{\textit{20}} & 3 & 16 & 1 & 0.50 & 0.7073 & 0.7058 & 0.6470 \\ \hline
\multicolumn{ 1}{c|}{\textbf{BELIEF+mCR *}} & \textbf{\textit{6}} & 1 & 0 & 5 & 0.17 & 0.7240 & 0.7788 & 0.6796 \\ 
\multicolumn{ 1}{c|}{} & \textbf{\textit{20}} & 4 & 10 & 6 & \textbf{0.67} & 0.6644 & 0.8006 & 0.6928 \\ \hline
\end{tabular}
}
\end{table}

\begin{table}[!htp]
\renewcommand{\arraystretch}{1.3}
\centering
\scriptsize
\caption{Evaluation results for Madelon. From left to right: the method name, the total number of features selected, and which and how many relevant, redundant and irrelevant are selected. The right-most part contains the prediction results using Na\"ive Bayes, Decision Tree, and Logistic Regression. Highlighted results in bold indicate the best outcome for each FS scheme. Rows with asterisk mean that a previous discretization phase is performed to enable NB prediction.}
\label{tab:madelon}
\resizebox{0.99\textwidth}{!}{
\begin{tabular}{c|r|c|c|c|c|r|r|r|r}
\hline
\multicolumn{1}{l|}{\textbf{}} & \multicolumn{ 6}{c|}{\textbf{Feature Selection}} & \multicolumn{ 3}{c}{\textbf{Accuracy}} \\ \hline
\multicolumn{1}{l|}{\textbf{Method}} & \multicolumn{1}{l|}{\textbf{\# sel.}} & \multicolumn{1}{l|}{\textbf{Rel.}} & \multicolumn{1}{l|}{\textbf{\# rel.}} & \multicolumn{1}{l|}{\textbf{\# red.}} & \multicolumn{1}{l|}{\textbf{\# irrel.}} & \multicolumn{1}{l|}{\textbf{success}} & \multicolumn{1}{l|}{\textbf{DT}} & \multicolumn{1}{l|}{\textbf{LR}} & \multicolumn{1}{l}{\textbf{NB}} \\ \hline
\multicolumn{ 1}{c|}{\textbf{BELIEF}} & \textbf{\textit{5}} & - & 0 & 0 & 5 & 0.00 & \textbf{\textit{0.6890}} & 0.6180 & \multicolumn{1}{c}{-} \\ 
\multicolumn{ 1}{c|}{} & \textbf{\textit{20}} & - & 0 & 0 & 20 & 0.00 & 0.7664 & 0.6045 & \multicolumn{1}{c}{-} \\ \hline
\multicolumn{ 1}{c|}{\textbf{BELIEF+mCR}} & \textbf{\textit{5}} & - & 0 & 0 & 5 & 0.00 & \textbf{\textit{0.6890}} & 0.6180 & \multicolumn{1}{c}{-} \\ 
\multicolumn{ 1}{c|}{} & \textbf{\textit{20}} & - & 0 & 0 & 20 & 0.00 & \textbf{\textit{0.7953}} & 0.6082 & \multicolumn{1}{c}{-} \\ \hline
\multicolumn{ 1}{c|}{\textbf{mRMR}} & \textbf{\textit{5}} & 1 & 1 & 0 & 4 & \textbf{0.20} & 0.6338 & 0.6083 & 0.6180 \\ 
\multicolumn{ 1}{c|}{} & \textbf{\textit{20}} & 1-5 & 5 & 5 & 10 & \textbf{1.00} & 0.6742 & 0.6101 & 0.6074 \\ \hline
\multicolumn{ 1}{c|}{\textbf{BELIEF *}} & \textbf{\textit{5}} & - & 0 & 0 & 5 & 0.00 & 0.6337 & 0.5761 & 0.5761 \\ 
\multicolumn{ 1}{c|}{} & \textbf{\textit{20}} & 1-2 & 2 & 0 & 18 & 0.40 & 0.6778 & 0.6128 & 0.6195 \\ \hline
\multicolumn{ 1}{c|}{\textbf{BELIEF+mCR *}} & \textbf{\textit{5}} & - & 0 & 0 & 5 & 0.00 & 0.6337 & 0.5761 & 0.5761 \\ 
\multicolumn{ 1}{c|}{} & \textbf{\textit{20}} & 1-5 & 5 & 0 & 15 & \textbf{1.00} & 0.6732 & 0.6177 & 0.6058 \\ \hline
\end{tabular}
}
\end{table}

\begin{figure*}[!htp]
\centering
\includegraphics[width=0.95\textwidth]{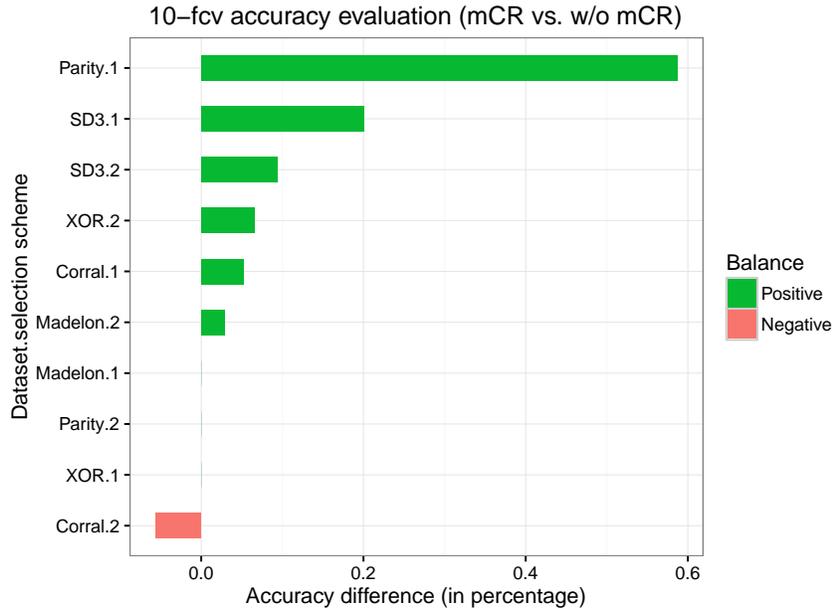}
\caption{10-fcv accuracy pairwise comparison BELIEF+mCR vs. BELIEF for several problems and selection schemes. Bars represent the difference between the best marks for each configuration. Row names are composed of the dataset name plus a number (FS configuration, 1: smaller selection, 2: large selection).}
\label{fig:acc-diff-mcr}
\end{figure*}

Furthermore, a pairwise comparison between the two BELIEF alternatives shows a great advantage on using mCR over the standard method. Figure~\ref{fig:acc-diff-mcr} clearly shows this advantage by depicting the difference between the best record achieved by each configuration. In 9/10 records mCR performs better or equal than the standard version. Indeed mCR is not only successful in redundant data ($ > 50\%$ leap in Parity-3+3), but also effective in redundancy-free problems (e.g.: XOR data).

What it is clear is that mCR perfectly fits its role of redundancy regulator as reflected in Tables~\ref{tab:parity},~\ref{tab:sd3}, and~\ref{tab:madelon}. From those tables we can notice the large number of redundant features embraced by standard BELIEF, and how mCR sharply downsizes this set. This fact is also reflected in the success formula, where mCR still performs better.

\subsection{BELIEF evaluation on large environments}

Once tested the robustness of mCR in small environments, we move to test its performance in large-scale scenarios. As a starting point, in previous sections we showed how mCR stands as a prominent alternative for small problems. Henceforth we aim at proving that BELIEF stands in large-scale FS as a competitive alternative when compared with the current state-of-the-art.

Starting from a small sampling rate, we provide fine-grained information about the discriminative power of BELIEF in supervised learning. Table~\ref{tab:acc-svc} shows F1 scores obtained by BELIEF, BELIEF+mCR and DmRMR in 1\% sampled. Note that with such small subgroup, BELIEF options are still able to overcome or at least not to be surpassed by DmRMR. Only in \textit{kddb}, DmRMR is better than BELIEF but with a difference $< 1\%$. This fact makes sense as high-dimensional problems have always been a stumble for NN-based algorithms. Conversely, a quantum leap can be noticed in \textit{ECBLD14} when applying the tuple BELIEF + mCR. 
 
\begin{table}[htbp]
\caption{F1-score results after selection and prediction (sampling rate: 0.01, classifier: SVC). In the first column, each block represents a dataset and different parameter configurations for BELIEF. Each of the other columns stand for a single valid FS combination, mixing different thresholds and algorithms. Beliefc indicates BELIEF + mCR. Best score by dataset and threshold is highlighted in bold, the overall best score by dataset is underlined, and the best BELIEF score is marked in italic.}
\begin{center}
\resizebox{1.\textwidth}{!}{
\begin{tabular}{lccccccccc}
\hline
\textbf{Method/Config.} & \multicolumn{1}{l}{\textbf{mRMR}} & \multicolumn{1}{l}{\textbf{Beliefc}} & \multicolumn{1}{l}{\textbf{Belief}} & \multicolumn{1}{l}{\textbf{mRMR}} & \multicolumn{1}{l}{\textbf{Beliefc}} & \multicolumn{1}{l}{\textbf{Belief}} & \multicolumn{1}{l}{\textbf{mRMR}} & \multicolumn{1}{l}{\textbf{Beliefc}} & \multicolumn{1}{l}{\textbf{Belief}} \\ 
 & \multicolumn{3}{c}{\textit{100 feat.}} & \multicolumn{3}{c}{\textit{50 feat.}} & \multicolumn{3}{c}{\textit{10 feat.}} \\ \hline
epsilon-k5-bs0.25-0 & \multicolumn{ 1}{c}{0.7910} & 0.6597 & 0.7190 & \multicolumn{ 1}{c}{0.7802} & 0.6175 & 0.6992 & \multicolumn{ 1}{c}{0.6961} & 0.5973 & 0.6331 \\ 
epsilon-k5-bs0.1-0 & \multicolumn{ 1}{c}{} & 0.7365 & 0.7491 & \multicolumn{ 1}{c}{} & 0.6909 & 0.7194 & \multicolumn{ 1}{c}{} & 0.6329 & 0.6647 \\ 
epsilon-k3-bs0.25-0 & \multicolumn{ 1}{c}{} & 0.6445 & 0.6761 & \multicolumn{ 1}{c}{} & 0.6149 & 0.6487 & \multicolumn{ 1}{c}{} & 0.5589 & 0.3338 \\ 
epsilon-k3-bs0.1-0 & \multicolumn{ 1}{c}{} & 0.6018 & 0.6625 & \multicolumn{ 1}{c}{} & 0.5872 & 0.6525 & \multicolumn{ 1}{c}{} & 0.5135 & 0.5052 \\ 
epsilon-k10-bs0.25-0 & \multicolumn{ 1}{c}{} & 0.7641 & 0.7956 & \multicolumn{ 1}{c}{} & 0.7247 & 0.7778 & \multicolumn{ 1}{c}{} & 0.6830 & 0.7050 \\ 
epsilon-k10-bs0.1-0 & \multicolumn{ 1}{c}{} & 0.7320 & \textbf{\textit{\underline{0.8027}}} & \multicolumn{ 1}{c}{} & 0.7214 & \textbf{0.7910} & \multicolumn{ 1}{c}{} & 0.6925 & \textbf{0.7067} \\  \hline
\textbf{baseline-0feat.} & 0.8976 &  & \textbf{} &  &  & \textbf{\textit{}} &  &  & \textbf{} \\ \hline
ecbdl-k5-bs0.25-0 & \multicolumn{ 1}{c}{0.4541} & \textbf{\textit{0.9749}} & 0.4599 & \multicolumn{ 1}{c}{0.4631} & 0.4729 & 0.4669 & \multicolumn{ 1}{c}{0.5047} & 0.6650 & 0.6802 \\ 
ecbdl-k5-bs0.1-0 & \multicolumn{ 1}{c}{} & \textbf{\textit{0.9749}} & 0.4612 & \multicolumn{ 1}{c}{} & 0.4761 & 0.4694 & \multicolumn{ 1}{c}{} & 0.6643 & 0.6802 \\ 
ecbdl-k3-bs0.25-0 & \multicolumn{ 1}{c}{} & 0.4565 & 0.4606 & \multicolumn{ 1}{c}{} & \textbf{0.4885} & 0.4787 & \multicolumn{ 1}{c}{} & 0.6650 & \textbf{0.6803} \\ 
ecbdl-k3-bs0.1-0 & \multicolumn{ 1}{c}{} & \textbf{\textit{\underline{0.9749}}} & 0.4606 & \multicolumn{ 1}{c}{} & 0.4756 & 0.4725 & \multicolumn{ 1}{c}{} & 0.6651 & 0.6802 \\ 
ecbdl-k10-bs0.25-0 & \multicolumn{ 1}{c}{} & 0.4521 & 0.4583 & \multicolumn{ 1}{c}{} & 0.4689 & 0.4709 & \multicolumn{ 1}{c}{} & 0.6643 & 0.6802 \\ 
ecbdl-k10-bs0.1-0 & \multicolumn{ 1}{c}{} & 0.4494 & 0.4496 & \multicolumn{ 1}{c}{} & 0.4675 & 0.4696 & \multicolumn{ 1}{c}{} & 0.6643 & 0.6643 \\ \hline
\textbf{baseline-0feat.} & 0.4376 &  &  &  &  &  &  &  &  \\ \hline
url-k5-bs0.25-0 & \multicolumn{ 1}{c}{\textbf{\underline{0.9640}}} & \textit{0.9611} & 0.9605 & \multicolumn{ 1}{c}{\textbf{0.9591}} & 0.9253 & 0.9263 & \multicolumn{ 1}{c}{\textbf{0.9448}} & 0.9270 & 0.9269 \\ 
url-k5-bs0.1-0 & \multicolumn{ 1}{c}{} & 0.9608 & 0.9603 & \multicolumn{ 1}{c}{} & 0.9258 & 0.9393 & \multicolumn{ 1}{c}{} & 0.9268 & 0.9274 \\ 
url-k3-bs0.25-0 & \multicolumn{ 1}{c}{} & 0.9610 & 0.9598 & \multicolumn{ 1}{c}{} & 0.9399 & 0.9397 & \multicolumn{ 1}{c}{} & 0.7016 & 0.5368 \\ 
url-k3-bs0.1-0 & \multicolumn{ 1}{c}{} & 0.9598 & 0.9603 & \multicolumn{ 1}{c}{} & 0.9412 & 0.9414 & \multicolumn{ 1}{c}{} & 0.5464 & 0.6869 \\ 
url-k10-bs0.25-0 & \multicolumn{ 1}{c}{} & 0.9605 & 0.9596 & \multicolumn{ 1}{c}{} & 0.9270 & 0.9274 & \multicolumn{ 1}{c}{} & 0.9268 & 0.9268 \\ 
url-k10-bs0.1-0 & \multicolumn{ 1}{c}{} & 0.9590 & 0.9600 & \multicolumn{ 1}{c}{} & 0.9257 & 0.9276 & \multicolumn{ 1}{c}{} & 0.9268 & 0.9268 \\ \hline
\textbf{baseline-0feat.} & 0.9876 & \multicolumn{1}{l}{} & \multicolumn{1}{l}{} & \multicolumn{1}{l}{} & \multicolumn{1}{l}{} & \multicolumn{1}{l}{} & \multicolumn{1}{l}{} & \multicolumn{1}{l}{} & \multicolumn{1}{l}{} \\ \hline
kddb-k5-bs0.25-0 & \multicolumn{ 1}{c}{\textbf{\underline{0.8413}}} & 0.8349 & 0.8349 & \multicolumn{ 1}{c}{0.8093} & 0.8327 & \textbf{0.8349} & \multicolumn{ 1}{c}{\textbf{0.8349}} & \textbf{\textit{0.8349}} & \textbf{\textit{0.8349}} \\ 
kddb-k5-bs0.1-0 & \multicolumn{ 1}{c}{} & 0.8343 & 0.8187 & \multicolumn{ 1}{c}{} & 0.8327 & 0.8327 & \multicolumn{ 1}{c}{} & \textbf{\textit{0.8349}} & \textbf{\textit{0.8349}} \\ 
kddb-k3-bs0.25-0 & \multicolumn{ 1}{c}{} & 0.8187 & 0.8349 & \multicolumn{ 1}{c}{} & \textbf{0.8349} & 0.8327 & \multicolumn{ 1}{c}{} & \textbf{\textit{0.8349}} & \textbf{\textit{0.8349}} \\ 
kddb-k3-bs0.1-0 & \multicolumn{ 1}{c}{} & 0.8349 & 0.8349 & \multicolumn{ 1}{c}{} & 0.8327 & 0.8155 & \multicolumn{ 1}{c}{} & \textbf{\textit{0.8349}} & \textbf{\textit{0.8349}} \\ 
kddb-k10-bs0.25-0 & \multicolumn{ 1}{c}{} & 0.8187 & 0.8187 & \multicolumn{ 1}{c}{} & 0.8327 & \textbf{0.8349} & \multicolumn{ 1}{c}{} & \textbf{\textit{0.8349}} & \textbf{\textit{0.8349}} \\ 
kddb-k10-bs0.1-0 & \multicolumn{ 1}{c}{} & 0.8349 & 0.8349 & \multicolumn{ 1}{c}{} & 0.8327 & 0.8327 & \multicolumn{ 1}{c}{} & \textbf{\textit{0.8349}} & \textbf{\textit{0.8349}} \\ \hline
\textbf{baseline-0feat.} & ? & \multicolumn{1}{l}{} & \multicolumn{1}{l}{} & \multicolumn{1}{l}{} & \multicolumn{1}{l}{} & \multicolumn{1}{l}{} & \multicolumn{1}{l}{} & \multicolumn{1}{l}{} & \multicolumn{1}{l}{} \\ \hline
\end{tabular}
}
\end{center}
\label{tab:acc-svc}
\end{table}

Focusing on BELIEF versions, we can assert that BELIEF+mCR comes to be more advantageous or similar in $3/4$ datasets. Its poor outcomes in~\textit{epsilon} may be caused by the high noisiness contained in this dataset. Additionally, redundancy among input features in \textit{epsilon} is negligible.

Table~\ref{tab:acc-dt} presents similar accuracy results but relying on tree learning to measure predictive power. Outcomes here seem not too illustrative since they tend to reproduce the same behavior shown in Table~\ref{tab:acc-svc}, but with lower scores.

%\begin{sidewaystable}
\begin{table}[htbp]
\caption{F1-score results after selection and prediction phases (sampling rate: 0.01, classifier: DT). In the first column, each block represents a dataset and different parameter configurations for BELIEF. Other columns mean a single combination, mixing different selection thresholds and algorithms. Best score by dataset and threshold is highlighted in bold, overall best score by dataset is underlined, and the best BELIEF score is marked in italic.}
\begin{center}
\resizebox{1.\textwidth}{!}{
\begin{tabular}{lccccccccc}
\hline
\textbf{Method/Config.} & \multicolumn{1}{l}{\textbf{mRMR}} & \multicolumn{1}{l}{\textbf{Beliefc}} & \multicolumn{1}{l}{\textbf{Belief}} & \multicolumn{1}{l}{\textbf{mRMR}} & \multicolumn{1}{l}{\textbf{Beliefc}} & \multicolumn{1}{l}{\textbf{Belief}} & \multicolumn{1}{l}{\textbf{mRMR}} & \multicolumn{1}{l}{\textbf{Beliefc}} & \multicolumn{1}{l}{\textbf{Belief-10}} \\
 & \multicolumn{3}{c}{\textit{100 feat.}} & \multicolumn{3}{c}{\textit{50 feat.}} & \multicolumn{3}{c}{\textit{10 feat.}} \\ \hline
epsilon-k5-bs0.25-0 & \textbf{\underline{0.6616}} & 0.6227 & 0.6516 & \textbf{0.6616} & 0.6081 & 0.6516 & \textbf{0.6545} & 0.6008 & 0.6269 \\ 
epsilon-k5-bs0.1-0 & \multicolumn{ 1}{c}{} & 0.6550 & 0.6595 & \multicolumn{ 1}{c}{} & 0.6391 & \textit{0.6595} & \multicolumn{ 1}{c}{} & 0.6335 & 0.6455 \\ 
epsilon-k3-bs0.25-0 & \multicolumn{ 1}{c}{} & 0.6062 & 0.6386 & \multicolumn{ 1}{c}{} & 0.5949 & 0.6341 & \multicolumn{ 1}{c}{} & 0.5591 & 0.4994 \\ 
epsilon-k3-bs0.1-0 & \multicolumn{ 1}{c}{} & 0.5771 & 0.6332 & \multicolumn{ 1}{c}{} & 0.5769 & 0.6297 & \multicolumn{ 1}{c}{} & 0.5129 & 0.4982 \\ 
epsilon-k10-bs0.25-0 & \multicolumn{ 1}{c}{} & 0.0000 & 0.0000 & \multicolumn{ 1}{c}{} & 0.0000 & 0.0000 & \multicolumn{ 1}{c}{} & 0.0000 & 0.0000 \\ 
epsilon-k10-bs0.1-0 & \multicolumn{ 1}{c}{} & 0.0000 & 0.0000 & \multicolumn{ 1}{c}{} & 0.0000 & 0.0000 & \multicolumn{ 1}{c}{} & 0.0000 & 0.0000 \\ \hline
\textbf{baseline-0feat.} & \multicolumn{ 1}{c}{0.6616} &  &  & \multicolumn{ 1}{c}{} &  &  & \multicolumn{ 1}{c}{} &  &  \\ \hline
ecbdl-k5-bs0.25-0 & 0.3864 & 0.4010 & 0.3481 & 0.3864 & 0.3994 & 0.3373 & 0.3225 & 0.4374 & 0.4732 \\ 
ecbdl-k5-bs0.1-0 & \multicolumn{ 1}{c}{} & 0.3577 & 0.4209 & \multicolumn{ 1}{c}{} & 0.4328 & 0.4136 & \multicolumn{ 1}{c}{} & 0.4366 & 0.4716 \\ 
ecbdl-k3-bs0.25-0 & \multicolumn{ 1}{c}{} & 0.3609 & 0.3511 & \multicolumn{ 1}{c}{} & \textbf{\textit{\underline{0.4798}}} & 0.3373 & \multicolumn{ 1}{c}{} & 0.4374 & \textbf{0.4779} \\ 
ecbdl-k3-bs0.1-0 & \multicolumn{ 1}{c}{} & 0.3978 & 0.3478 & \multicolumn{ 1}{c}{} & 0.4327 & 0.3355 & \multicolumn{ 1}{c}{} & 0.4373 & 0.4729 \\ 
ecbdl-k10-bs0.25-0 & \multicolumn{ 1}{c}{} & \textbf{0.4379} & 0.3608 & \multicolumn{ 1}{c}{} & 0.4368 & 0.3566 & \multicolumn{ 1}{c}{} & 0.4368 & 0.4716 \\ 
ecbdl-k10-bs0.1-0 & \multicolumn{ 1}{c}{} & 0.3800 & 0.4070 & \multicolumn{ 1}{c}{} & 0.3458 & 0.4178 & \multicolumn{ 1}{c}{} & 0.4388 & 0.4388 \\ \hline
\textbf{baseline-0feat.} & \multicolumn{ 1}{c}{0.3789} &  &  & \multicolumn{ 1}{c}{} &  &  & \multicolumn{ 1}{c}{} &  &  \\ \hline
url-k5-bs0.25-0 & \textbf{\underline{0.9635}} & 0.9573 & 0.9573 & 0.9586 & 0.9587 & 0.9424 & 0.9568 & 0.9375 & 0.9375 \\ 
url-k5-bs0.1-0 & \multicolumn{ 1}{c}{} & 0.9574 & 0.9574 & \multicolumn{ 1}{c}{} & 0.9586 & 0.9425 & \multicolumn{ 1}{c}{} & 0.9375 & 0.9375 \\ 
url-k3-bs0.25-0 & \multicolumn{ 1}{c}{} & 0.9573 & 0.9573 & \multicolumn{ 1}{c}{} & 0.9490 & 0.9490 & \multicolumn{ 1}{c}{} & 0.7671 & 0.8075 \\ 
url-k3-bs0.1-0 & \multicolumn{ 1}{c}{} & 0.9577 & 0.9577 & \multicolumn{ 1}{c}{} & 0.9490 & 0.9490 & \multicolumn{ 1}{c}{} & 0.8073 & 0.8055 \\ 
url-k10-bs0.25-0 & \multicolumn{ 1}{c}{} & 0.9573 & 0.9573 & \multicolumn{ 1}{c}{} & 0.9428 & 0.9457 & \multicolumn{ 1}{c}{} & 0.9375 & 0.9373 \\ 
url-k10-bs0.1-0 & \multicolumn{ 1}{c}{} & 0.9577 & 0.9577 & \multicolumn{ 1}{c}{} & \textbf{\textit{0.9608}} & 0.9425 & \multicolumn{ 1}{c}{} & \textbf{0.9395} & \textbf{0.9395} \\ \hline
\textbf{baseline-0feat.} & \multicolumn{ 1}{c}{?} & \multicolumn{1}{l}{} & \multicolumn{1}{l}{} & \multicolumn{ 1}{l}{} & \multicolumn{1}{l}{} & \multicolumn{1}{l}{} & \multicolumn{ 1}{l}{} & \multicolumn{1}{l}{} & \multicolumn{1}{l}{} \\ \hline
kddb-k5-bs0.25-0 & \textbf{\underline{0.8376}} & 0.8349 & 0.8349 & \textbf{0.8372} & 0.8349 & 0.8349 & \textbf{0.8367} & \textit{0.8349} & \textit{0.8349} \\ 
kddb-k5-bs0.1-0 & \multicolumn{ 1}{c}{} & 0.8349 & 0.8349 & \multicolumn{ 1}{c}{} & 0.8349 & 0.8349 & \multicolumn{ 1}{c}{} & \textit{0.8349} & \textit{0.8349} \\ 
kddb-k3-bs0.25-0 & \multicolumn{ 1}{c}{} & 0.8349 & 0.8349 & \multicolumn{ 1}{c}{} & 0.8349 & 0.8349 & \multicolumn{ 1}{c}{} & \textit{0.8349} & \textit{0.8349} \\ 
kddb-k3-bs0.1-0 & \multicolumn{ 1}{c}{} & 0.8349 & 0.8349 & \multicolumn{ 1}{c}{} & 0.8349 & 0.8349 & \multicolumn{ 1}{c}{} & \textit{0.8349} & \textit{0.8349} \\ 
kddb-k10-bs0.25-0 & \multicolumn{ 1}{c}{} & 0.8349 & 0.8349 & \multicolumn{ 1}{c}{} & 0.8349 & 0.8349 & \multicolumn{ 1}{c}{} & \textit{0.8349} & \textit{0.8349} \\ 
kddb-k10-bs0.1-0 & \multicolumn{ 1}{c}{} & 0.8349 & 0.8349 & \multicolumn{ 1}{c}{} & 0.8349 & 0.8349 & \multicolumn{ 1}{c}{} & \textit{0.8349} & \textit{0.8349} \\ \hline 
\textbf{baseline-0feat.} & \multicolumn{ 1}{c}{?} & \multicolumn{1}{l}{} & \multicolumn{1}{l}{} & \multicolumn{ 1}{l}{} & \multicolumn{1}{l}{} & \multicolumn{1}{l}{} & \multicolumn{ 1}{l}{} & \multicolumn{1}{l}{} & \multicolumn{1}{l}{} \\ \hline
\end{tabular}
}
\end{center}
\label{tab:acc-dt}
\end{table}

\subsection{Sampling rate impact on BELIEF}
\label{subsec:sampling-impact}

\begin{figure}[!thp]
  \begin{center}
   \includegraphics[width=.8\textwidth]{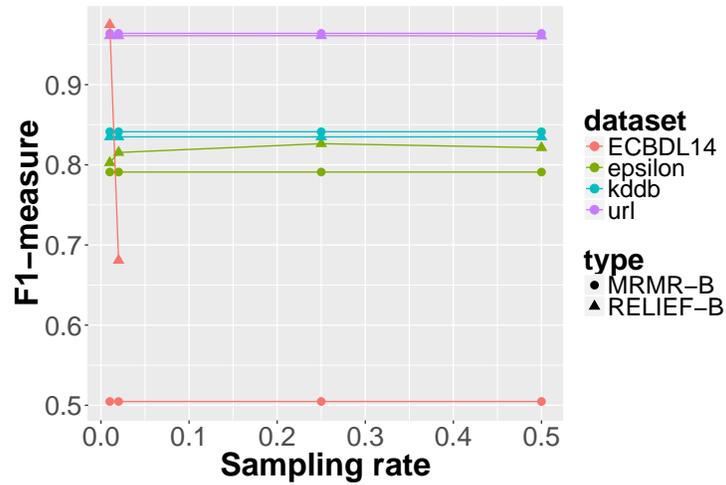}
  \caption{Evolution of F1 measure as sampling rate is increased. For each dataset the most accurate record is depicted. MRMR-B stands for DmRMR with 100 features selected.}
  \label{fig:f1-macc}
  \end{center}
\end{figure}

\begin{figure}[!thp]
  \begin{center}
   \includegraphics[width=.8\textwidth]{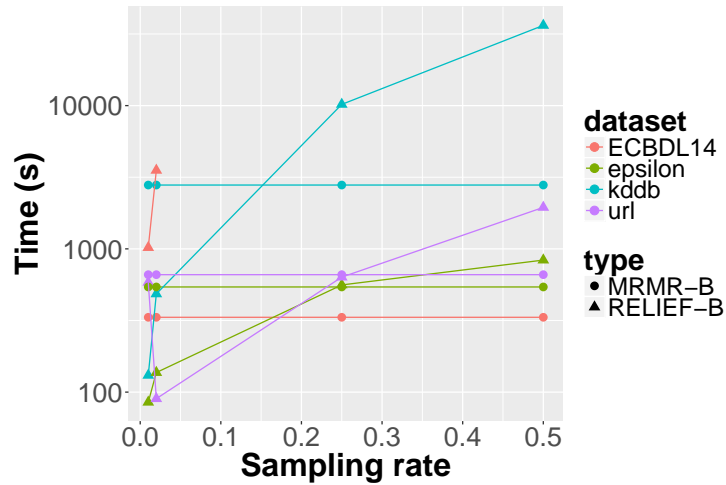}
  \caption{Evolution of runtime performance as sampling rate is increased. For each dataset the most accurate record is depicted. MRMR-B stands for DmRMR with 100 features selected.}
  \label{fig:time-macc}
  \end{center}
\end{figure}

In this section we inspect precision and runtime performance as sampling rate augments, as well as, to analyze the possible negative impact of an excessively small rate. Figures~\ref{fig:f1-macc} and~\ref{fig:time-macc} contain information about the impact of sampling size on accuracy (F1-score). In Figure~\ref{fig:f1-macc}, we can observe the stability of scores as sampling augments. According to the graph, only $1\%$ of data is enough to properly estimate weights in most of cases (except in epsilon). Time measurements recorded in Figure~\ref{fig:time-macc} also confirm low rates in BELIEF provides lower reaction times compared to DmRMR.

\begin{figure}[!thp]
  \begin{center}
   \includegraphics[width=.8\textwidth]{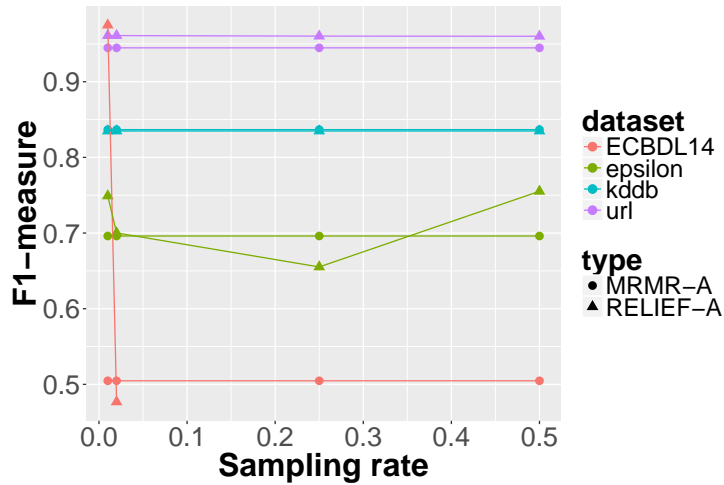}
  \caption{Evolution of F1 measure as sampling rate is increased. For each dataset the most rapid record is depicted. MRMR-A stands for DmRMR with 10 features selected.}
  \label{fig:f1-mrapid}
  \end{center}
\end{figure}

\begin{figure}[!thp]
  \begin{center}
   \includegraphics[width=.8\textwidth]{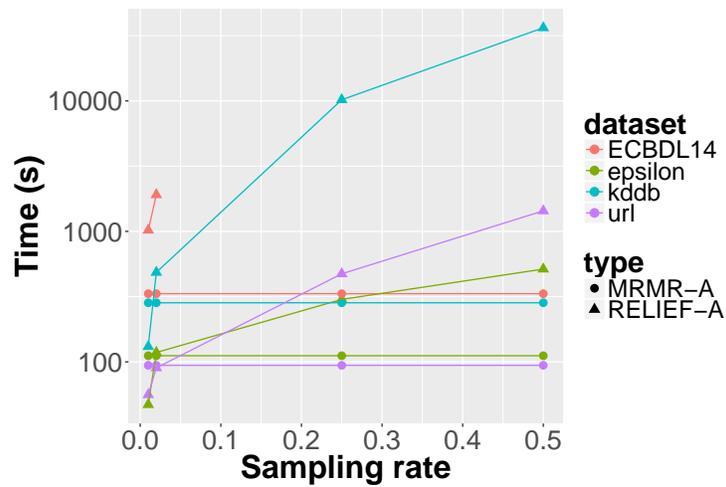}
  \caption{Evolution of runtime performance as sampling rate is increased. For each dataset the most rapid record is depicted. MRMR-A stands for DmRMR with 10 features selected.}
  \label{fig:time-mrapid}
  \end{center}
\end{figure}

In case we rather focus on rapid solutions, Figures~\ref{fig:f1-mrapid} and~\ref{fig:time-mrapid} depict empirical results for the most agile configurations in BELIEF and DmRMR. According to these results, we can underpin BELIEF again as the most effective and efficient alternative for low-rate scenarios. For 1\% sampled data, BELIEF obtains substantial benefits with respect to DmRMR regarding time performance. 
%In this case BELIEF's schemes are even more precise for \textit{url} and \textit{epsilon} datasets. 
Beyond a higher accuracy, low-rate solutions plus offer faster response times than DmRMR. Higher rates seems to be not interesting because of their high runtime cost and lack of accuracy improvement.

In order to study the impact of sampling over-reduction on feature weighting, we have compared the precision results obtained by our smallest scheme (1\% of sampling) against those proposed in DiReliefF~\cite{palma18}, with only 100 instances. %From this plot we can draw some conclusions about the negative effects of relying on tiny samples. 
Results on both datasets show a clear drop on F1 score when using extremely small rates, such as only 100 instances from datasets with millions. Then, we can assert with some certainty proper sampling rates render as essential to obtain fair estimations. A solution that allows scaling of sampling set is thus required.

\begin{figure}[!thp]
  \begin{center}
   \includegraphics[width=.8\textwidth]{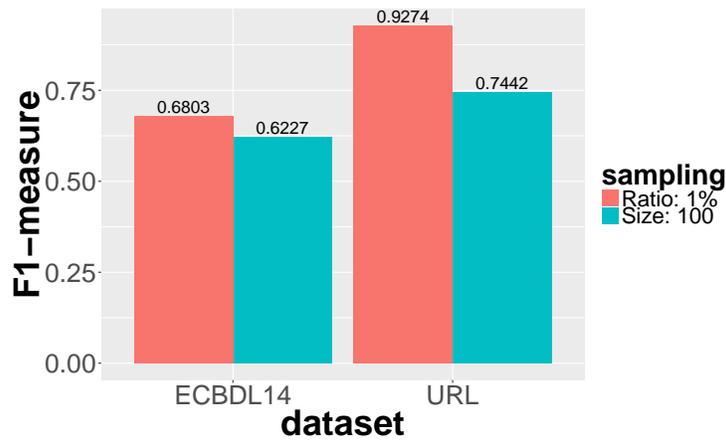}
  \caption{Impact on F1 score associated to over-reduction on sampling rate: 100 instances sampled vs. 1\% of sampling rate (about thousands). One dataset for each type was selected (sparse and dense). And the selection number was bounded to the most demanding configuration, namely, only 10 features.}
  \label{fig:overreduction}
  \end{center}
\end{figure}

\section{Concluding Remarks}
\label{sec:conc}

In this paper, we have presented BELIEF, a feature weighting algorithm capable of accurately estimating feature importance in large samples --both in number of features and examples--. With this new proposal we aimed at solving the performance deficiencies shown by RELIEF and its distributed versions when facing big datasets, concretely, large estimation samples. By restricting network communication among partitions, and the scope of main computations (moving from instance-wise to partition-wise) in RELIEF, we have managed to create a proper distributed version able to naturally scale up as requested.

Aside from the fully-scalable model, we have addressed lack of redundancy management techniques in RELIEF models by proposing a built-in redundancy removal procedure that detects and eliminates duplicities in selections. Our solution (mCR) relies on feature co-occurrences (collisions) already computed to estimate strong dependencies among input features at barely no time cost. Experiments performed on small data confirmed that no substantial difference between schemes generated by mCR and other information-based models exists beyond a higher cost imposed by the latter ones. 

Extended tests comprising several real-world datasets --up to $O(10^7)$ instances and $O(10^4)$ features-- have asserted as well the relevance of BELIEF in large environments. Results shed light about BELIEF's predominance in terms of runtime performance and precision of schemes when compared to alternative models such as DiReliefF. As future work, we plan to incorporate some mechanisms to further expedite neighbor searches in BELIEF. They will be based on locality sensitivity hashing tables, or metric trees.

\section*{Acknowledgements}
This work is supported by the Spanish National Research Project TIN2014-57251-P, the Foundation BBVA project 75/2016 BigDaP-TOOLS - ``Ayudas Fundaci\'on BBVA a Equipos de Investigaci\'on Cient\'ifica 2016", the Andalusian Research Plan P11-TIC-7765. S. Ram\'irez-Gallego holds a FPU scholarship from the Spanish Ministry of Education and Science (FPU13/00047).

\bibliography{biblio}

\end{document}